\documentclass[conference,10pt]{IEEEtran}
\IEEEoverridecommandlockouts
\usepackage{cite}
\usepackage{amsmath,amssymb,amsfonts}
\usepackage{graphicx}
\usepackage{textcomp}
\usepackage{xcolor}
\usepackage{subfigure}
\usepackage{epstopdf}
\usepackage{epsfig}
\def\BibTeX{{\rm B\kern-.05em{\sc i\kern-.025em b}\kern-.08em
		T\kern-.1667em\lower.7ex\hbox{E}\kern-.125emX}}
\usepackage{algpseudocode}
\usepackage[ruled,vlined]{algorithm2e}
\begin{document}

\title{MAT-CNN-SOPC: Motionless Analysis of Traffic Using Convolutional Neural Networks on System-On-a-Programmable-Chip\\
\thanks{This work is supported by the UK Engineering and Physical Sciences Research Council EPSRC [EP/R02572X/1 and EP/P017487/1].\newline \newline
	978-1-5386-7753-7/18/\$31.00 ©2018 IEEE}
}


\author{\IEEEauthorblockN{Somdip Dey, Grigorios Kalliatakis, Sangeet Saha, Amit Kumar Singh, Shoaib Ehsan, Klaus McDonald-Maier}
\IEEEauthorblockA{Embedded and Intelligent Systems Laboratory\\
\textit{University of Essex}\\
Colchester, UK \\
\{~somdip.dey, gkallia, sangeet.saha, a.k.singh, 
sehsan, kdm \} @essex.ac.uk
}}

\maketitle

\begin{abstract}
Intelligent Transportation Systems (ITS) have become an important pillar in modern ``smart city'' framework which demands intelligent involvement of machines. Traffic load recognition can be categorized as an important and challenging issue for such systems. Recently, Convolutional Neural Network (CNN) models have drawn considerable amount of interest in many areas such as weather classification, human rights violation detection through images, due to its accurate prediction capabilities. This work tackles real-life traffic load recognition problem on System-On-a-Programmable-Chip (SOPC) platform and coin it as MAT-CNN-SOPC, which uses an intelligent re-training mechanism of the CNN with known environments. The proposed methodology is capable of enhancing the efficacy of the approach by 2.44x in comparison to the state-of-art and proven through experimental analysis. We have also introduced a mathematical equation, which is capable of quantifying the suitability of using different CNN models over the other for a particular application based implementation.
\end{abstract}


\section{Introduction}
Some of the popular ways of traffic monitoring and analysis for categorization of traffic load is either using vehicle based assess method \cite{beymer1997real, chen2012traffic, jung2001content, andrews2013highway, asmaa2013road, hu2012real, riaz2013traffic, luo2015traffic} or a holistic approach \cite{chan2005classification, derpanis2011classification, porikli2004traffic, ess2009segmentation}. But analysis of traffic using these popular methods require high frame rate videos with a stable environmental condition, which could be the biggest limiting factor in many places. Without these conditions being met \cite{kastrinaki2003survey, luo2015traffic, luo2016traffic, wang2017interactive}, reliable motion features cannot be extracted, which might  result in corrupted output.

Because of large-scale camera network not being able to stream and store high-frame rate videos gathered by a network of interconnected cameras due to bandwidth limitation and limited on-board storage capacity, streaming low-frame videos on these camera is very common. In many cases when these cameras stream over a WIFI network, it is often difficult to stream more than 2 frames per second due to the limited bandwidth of the network \cite{luo2015traffic, luo2016traffic}. Moreover due to cost constraint of such interconnected camera networks and associated servers, many developing countries might not be able to adopt and implement such sophisticated state-of-the-art traffic analysis and categorization methodologies. On the other hand image processing \cite{jain1995machine, dey2012amalgamation, dey2013sd, dey2012image, dey2012advanced, dey2013confidential, dey2012modern, dey2012advanced2} and computer vision applications \cite{jain1995machine, ehsan2015integral, viola2004robust} are very well known for their thread, task and data level parallelism. Recently we could also notice a huge increase in integrating Convolutional Neural Networks \cite{chakradhar2010dynamically, chen2014big, vgg2014, alexnet2012, resnet2016} in computer vision to solve several real-life challenges such as human rights violation detection through images \cite{kalliatakis2017paradigm, kalliatakis2017detection}, weather forecasting \cite{zhang1998forecasting, grecu1999detection}, etc. Due to high level of data parallelism in computer vision applications using Convolutional Neural Networks and reducing cost factor of field-programmable gate array (FPGA) based system-on-a-programmable-chip (SOPC) \cite{intelFPGA, bdtiFPGA}, such SOPC serves as a cost-effective option to analyze and categorize traffic.

In this paper, we propose a novel methodology to analyze and categorize traffic using Convolutional Neural Networks on SOPC without the need of streaming the video-frames to the server for further categorization as is usually done in state-of-the-art traffic categorization methodologies. The proposed methodology is coined as Motionless Analysis of Traffic Using Convolutional Neural Networks on SOPC: MAT-CNN-SOPC and we have also introduced a \textit{Quality of Experience} variable, which would enhance the predicting mechanism of the chosen CNN model. The remainder of this paper is organized as follows. Section \ref{relatedwork} mentions the related work in the field and Section \ref{sysproblemformulation} provides a breakdown of the software and hardware infrastructure used for the implementation and validation purposes of the proposed methodology along with the dataset used and problem definition solved using this solution. Section \ref{propoedMethod} provides a comprehensive view of the proposed methodology and in Section \ref{results} we could analyze the experimental results. Section \ref{discuss} briefly mentions some related discussion on the proposed methodology. Finally, Section \ref{conclusion} concludes the paper.

\section{Related Work}\label{relatedwork}
Before 2015, majority of traffic analysis and categorization was mostly performed using the following methodologies:

\begin{itemize}
	\item Vehicle based methodologies where either vehicles are first localized on the road with a background subtraction method \cite{jung2001content, andrews2013highway, asmaa2013road} or the vehicles are localized with moving feature keypoints \cite{ hu2012real, riaz2013traffic}. In these methodologies the resulting tracks are concatenated together to identify key features of traffic such as traffic lanes, average traffic speed, average traffic density, etc.
	\item A holistic approach, where a macroscopic analysis of traffic flow is understood through global representation of a scene, which is obtained by accounting for spatio-temporal features except tracking using background subtraction and moving feature keypoints \cite{chan2005classification, porikli2004traffic, derpanis2011classification}.
\end{itemize}

Although the aforementioned methodologies are highly effective to analyze traffic, the biggest limiting factor is the cost of sophisticated camera-network involved and the requirement for high-frame-rate videos to compute reliable motion features. To break away from this trend of traffic analysis, in 2015 Luo et al. \cite{luo2015traffic} proposed a methodology to use various image processing and CNN based approaches to analyze traffic without moving features. In this paper the authors used four different visual descriptors such as bag of visual words (BOVW), Vector of Locally Aggregated Descriptors (VLAD), improved Fisher Vector (IFV) and Locality-constrained Linear Coding (LLC), and have also used pre-trained deep CNN models such as Caffe and VGG to analyze traffic and predict categorization of the same. The approach taken by Luo et al. to use popular image processing and CNN methods to classify traffic is novel and solves the low-frame-rate video streaming issue. However, the experimental setups and results provided in the paper is susceptible to some biasness as the cross-dataset validation was not performed. In Section \ref{discuss} we have compared our experimental setup and achieved results with the ones mentioned in \cite{luo2015traffic}. In another extended paper published by Luo et al. \cite{luo2016traffic}, the researchers have used SegCNN and RegCNN to analyze and classify traffic. In both the aforementioned papers the authors are training and classifying traffic images after the video frames are transferred to the server from the interconnected camera network. But installing and implementing such hardware infrastructure to analyze traffic in developing countries is a challenging issue\cite{mokhaareview}.

Other state-of-the-art methodologies include detecting \& counting the numbers of cars and computing traffic density based on that using CNN-based vehicle detectors with high accuracy at near real time \cite{zhang2015integrated, cai2016unified, zhou2016image}. Although this way of detecting traffic density could still be classified as a vehicle based approach and has become popular in recent times but there are associated challenges with these methods as follows:
\begin{itemize}
	\item Training and test data should belong to the same dataset taken from the same camera with same configuration and hence require consistency in training.
	\item Cars detected need to be within a particular range or scope of the image and these methodologies fail to detect cars, which are far away in the images captured.
	\item These methodologies performed poorly if the captured images were occluded, especially in case of heavy traffic \& jam.
\end{itemize}

From the aforementioned list of issues with the state-of-the-art methods, although Deep Learning \cite{lecun2015deep} could  solve the problem of detecting occluded objects properly but such method usually requires large dataset to be trained with. But for the application of traffic categorization there is no such publicly available dataset and hence using Deep Learning would be inefficient.

Compared to all the aforementioned works, we propose an easy to train CNN model, which do not require a lot of images in the training dataset, with combination of transfer learning \footnote{Learning achieved by taking the convolutional base of a pre-trained network, running the new data of 4 traffic categories through it and training a new randomly initialized classifier} and continuous learning \footnote{Learning achieved by re-training the classifier with wrong predictions till operating period of the system} capabilities on SOPC without the need of communicating the traffic images to the connected server for further analysis.

\section{System and Problem Formulation}\label{sysproblemformulation}
\subsection{Hardware Infrastructure \& Software Infrastructure}\label{hardwaresoftware}
It is worth mentioning that the CNN based traffic analysis will demand a huge amount of computing resources. Rather than high performance general purpose processing unit, the application specific computing could also be a lucrative way out. From the recent literature studies \cite{lecun2015deep, transferLearning2010}, it has been observed that software based execution could provide the required flexibility but not the performance efficiency in terms of execution. On the  other hand, a dedicated hardware based execution will provide performance efficacy but will under perform when the flexibility becomes the major concern. 

Thus, hardware software co-execution ecosystem is emerging as a bright prospect and modern FPGA (ZYNQ) platform is a good solution to implement such functionality \footnote{Even though GPUs could be an efficient  accelerators for CNNs. However, such devices are expensive \& very power hungry and thus, make them not suitable in the aforementioned power-constrained scenarios}. In order to carry out the functionality in FPGA, we have chosen the vivado HLS \cite{vivado} framework. This framework also extracts the parallelism inside the code. The entire CNN model is created in high level language (C/C++, Matlab, Python). Then it has been converted in to RTL \footnote{~RTL: Register-transfer level is a design abstraction, which models a synchronous digital circuit in terms of the flow of digital data between hardware registers and the logical operations performed on those signals.} format through vivado high level synthesis. Once the code has been converted, the VIVADO framework will synthesize to the bitstream to make the design executable. Our code (in Matlab, Python \& C/C++) is provided on our GitHub repository \cite{gitHubRepo}.

\subsection{Dataset}\label{datasetUsed}
For our research we are using the same dataset used by Luo et al. \cite{luo2015traffic, luo2016traffic} to validate performance of our proposed methodology and theories. Mainly two dataset are used. The first one is the dataset released by UCSD traffic control department \cite{datasetChan2005classification}. This dataset contains 254 highway video sequences, all filmed by the same camera containing light, heavy and traffic jams filmed at different periods of the day and under different weather conditions. Each UCSD video has a resolution of 320 X 240 with a frame rate of 10 fps.

\begin{figure}[htbp]
	\centerline{\includegraphics[width=7cm,height=7cm,keepaspectratio]{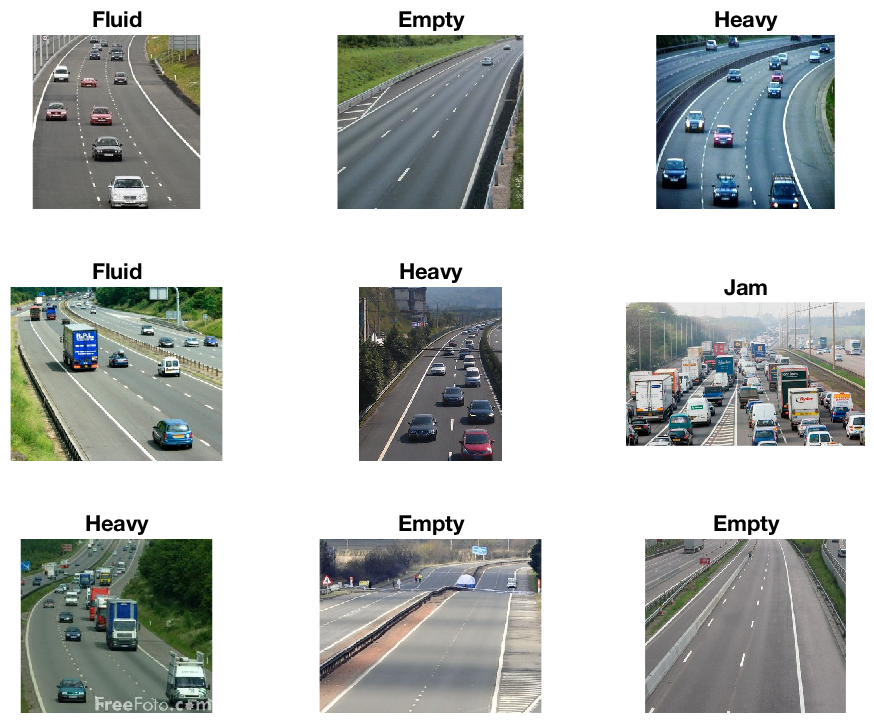}}
	\caption{Random images from 4 Categories of Traffic Classification: Jam, Heavy, Fluid, Empty \cite{luo2015traffic}}
	\label{trafficClassFig}
\end{figure}

The second dataset consist of the 400 images\footnote{Only 400 images were available in the existing dataset provided by Luo et al.~\cite{luo2015traffic}} captured from highway cameras deployed in all over the UK and also consist of several examples of different weather and lighting conditions in order to provide a better training performance. These 400 images are segregated into 4 categories: Jam, Heavy, Fluid, Empty (as shown in Fig. \ref{trafficClassFig}), and each category having 100 images.

\subsection{Problem Definition}
The main focus of this research is to be able to implement a hardware-software ecosystem, which is able to analyze and predict traffic effectively on the System-on-programmable-chip without streaming the video-frames to the server over a communication channel even in severe hardware impaired conditions such as poor video recording capabilities of the camera. Since a practical application such as categorization of traffic using CNNs methodologies requires a desirable ``Quality of Experience" (\textit{QoE}) in order to be a successful implementation, we also need to define the governing equation to quantify \textit{QoE} so that we could understand the overall desirability of the CovNet methodology being used for the problem in hand. Let us consider the (\textit{QoE}) that will decide whether the accuracy of the CovNet methodology is desirable as \textit{Q} and the predicted label (categorization) of the CovNet as P\textsubscript{i} for any image (\textit{i}) from a dataset of images (\textit{I}) at an instance. Then the governing equation which could be used to predict the label (category) of the traffic as desirable at an instance is as follows:

\begin{equation}\label{eqQoE}
\forall \{i \in I : i > 1\}, P_{i} \geq Q
\end{equation}

In the aforementioned equation (\ref{eqQoE}), we are not taking the training time of the CovNet model into consideration as part of \textit{QoE} since it is assumed that training is mandatory and completed while the hardware-software ecosystem is setup on the section of the road or highway for the purpose of categorizing traffic. Later in Section \ref{discuss} we would also provide a minimum threshold value for \textit{QoE} for the given problem in hand based on the experimental results (Section \ref{results}) performed.

\section{Proposed Methodology: MAT-CNN-SOPC}\label{propoedMethod}

 \begin{figure}[!t]
 		\centerline{\includegraphics[scale=0.30]{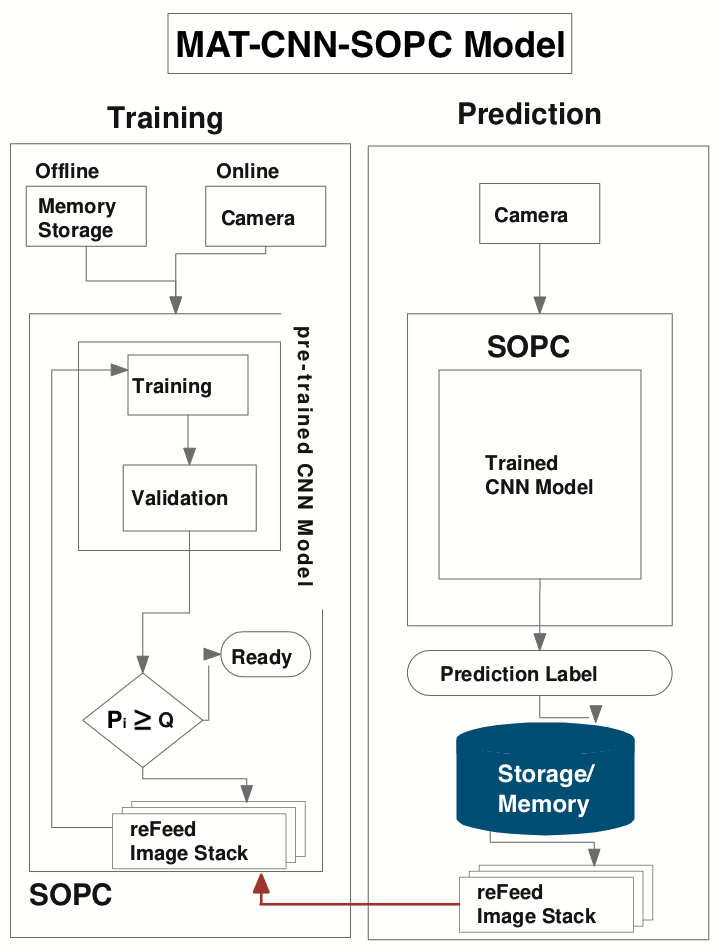}}
 		\caption{MAT-CNN-SOPC Model  Work flow}
 		\label{mainModel}
	\centering
\end{figure}

In this section we propose the hardware-software ecosystem, MAT-CNN-SOPC, which would be utilizing the categorization power of a pre-trained CNN model to be trained to effectively categorize traffic based on the desired categories. We propose a two fold module of MAT-CNN-SOPC: \textit{Training} \& \textit{Prediction} (as shown in Fig. \ref{mainModel}). Both the \textit{Training} and \textit{Prediction} modules are implemented in application layer of the SOPC. For this hardware-software ecosystem we assume that a camera is connected to the system-on-a-programmable-chip and the primary training of the classifier of the pre-trained CNN model is performed \footnote{Using transfer learning of pre-trained CNN model} while the SOPC is setup on the section of the road in the first place. 

For our proposed model we could select any available pre-trained CNN model such as AlexNet \cite{alexnet2012}, VGG \cite{vgg2014}, ResNet \cite{resnet2016}, etc. for the \textit{Training} module. In this module we train the system with various known images of traffic. Since FPGA on the SOPC are excellent candidates for SIMD programming exploration, we use FPGA on board as accelerators for the Convolutional layers during the training. The training module consists of both offline training as well as online training. During the offline training, the model is trained on the dataset, which is either pre-stored on the SOPC or stored on an external storage connected to the system. After the initial (offline) training is complete with the pre-stored dataset,  the camera connected on the SOPC is activated to send in images of the current traffic/section of the road with determined labels (categories) and the training of the model is validated. If the model predicts a wrong category of the streamed image then that image along with it's correct category is stored in a \textit{reFeed Image Stack}, a special stack implementation to hold images with labels, on the system for later (online) training. If during this validation stage of the model, the total prediction accuracy falls below the desired accuracy (\textit{Q} as mentioned in Eq. \ref{eqQoE}) of the model then the model is re-trained with the images stored in the \textit{reFeed Image Stack}. After completion of every training process the validation phase is re-executed till the prediction accuracy of the model is equal or more than \textit{Q} (Quality of Experience). Methodology of the training module is algorithmically provided in Algo. \ref{algo1}. The main motivation to re-train the CNN model with failed prediction dataset of a known environment is to artificially enhance the accuracy of the model and we call this enrichment in performance as \textit{reFeed Gain factor (r)}. In Section \ref{testResultTab}, we have provided the value of \textit{reFeed Gain factor}: \textit{r} noted from the performed experiments and we have also provided a generic mathematical notation of this terminology for better representation as follows:

\begin{equation}\label{eqrFG}
r =  |P_{i}^{f} - P_{i}^{0}|, \text{where} ~P_{i}^{0} \leq Q \leq P_{i}^{f} 
\end{equation}

In the aforementioned equation (Eq. \ref{eqrFG}) \textit{Q} is the Quality of Experience (see Eq. \ref{eqQoE}), which is desired for the system to perform well (related to predicting traffic categories),  $P_{i}^{f}$ is the prediction accuracy of the CNN model after re-trained with \textit{reFeed Image Stack} and $P_{i}^{0}$ is the prediction accuracy in the initial training.

Based on Eq. \ref{eqrFG}, if we consider \textit{S} as the boost function in prediction accuracy of the CNN model after re-training with reFeed Image Stack feature, which we denote as \textit{reFeed Gain} (\textit{R}), we could represent the \textit{reFeed Gain} as follows:

\begin{equation}\label{eqR}
R \leftarrow S(P_{i}^{0}) = (P_{i}^{f} / P_{i}^{0})
\end{equation}

Therefore, using Eq. \ref{eqrFG} \& \ref{eqR} we could generalize the relationship between $P_{i}^{0}$, $R$, $r$, $Q$ ~\footnote{~$P_{i}^{0}$ denotes initial prediction accuracy, $R$ denotes reFeed Gain, $r$ denotes reFeed Gain Factor and $Q$ corresponds to the Quality of Experience} as follows:

\begin{equation}\label{eqRRelationship}
R \times P_{i}^{0} = r + P_{i}^{0},~ \text{where} ~P_{i}^{0} \leq Q
\end{equation}

Now, in the prediction module our CNN model keeps predicting the traffic category (label) and it either broadcasts the label over the network or it stores the labels along with the video frames on a memory storage, which could be either on-board or external. Later we could use the concept of ``\textit{assistive learning}", where a human being manually goes through the stored video frames along with their predicted labels and rectifies any label if there was a wrong prediction. Whenever an image is classified as wrong by the assistive human being then that image goes into the \textit{reFeed Image Stack} of the Prediction module and later the images from this stack is transfered to the \textit{reFeed Image Stack} in the Training module so that the CNN model could be further trained with the images from the \textit{reFeed Image Stack} to enhance \textit{reFeed Gain} ($R$). We call this method to improving the prediction accuracy of the existing CNN model as ``\textit{Continuous Learning}" of the CNN Model for a specific category (as shown in Fig. \ref{mainModel}). In this particular work we are only focused on the implementation of \textit{reFeed Image Stack} and \textit{reFeed Gain} in the Training module.

\begin{algorithm} [!h]
	\small{
		\KwIn{\\1. $\mathcal{I}$: set of $n$ Images from Training \& Validation Dataset\\
		2. $\mathcal{T}$: set of $m$ Images from Testing Dataset (for cross-validation)}
		\KwOut{$P$: prediction accuracy after training }
		
		\textbf{Initialize:} $\mathcal{Q}$ = 0.7;     \Comment{Quality of Experience is set to 70\% by default}\\
		$\mathcal{S}.Count$ = 0; \Comment{$\mathcal{S}$: \textit{reFeed Image Stack}}\\
		 \smallskip
		\underline{\textbf{Offline Training:}}\\
		Train~(pre-trained CNN model~, ~$\mathcal{I}$); \Comment{Train model with $\mathcal{I}$ dataset}\\
		\For{each image $i \in \mathcal{T}$}
		{Prediction = Test~(~CNN model~); \Comment{Test outputs whether prediction is correct or wrong}\\
			\Comment{Prediction.IsWrong() is a function to return True when Prediction.Label $!=$ Original.Label of Test image $i$}\\
			\If{Prediction.IsWrong()}{$S.Push(i)$;}
			Calculate mean Prediction Accuracy ($P_{i}^{0}$);
		}
		$P = P_{i}^{0}$;\\
		\smallskip
		\underline{\textbf{Online Training:}}\\
		\{\textit{re-Train with reFeed Image Stack if $P_{i}^{0} < Q$}\}\\
		\If{$P_{i}^{0} < Q$}{\Comment{Need to satisfy condition of Eq. \ref{eqQoE}}\\
			\If{S.Empty() == False}{
					\{\textit{Traing CNN with \textit{reFeed Image Stack}}\}\\
					Train~(CNN model~,~~$\mathcal{S}$ );\\
					Calculate mean Prediction Accuracy ($P_{i}^{f}$);
			}
			$P = P_{i}^{f}$;\\
			$\mathcal{S}.Count$ = 0; \Comment{reset \textit{reFeed Image Stack}}
		}
		\Else{return $P$;}
		
		\caption{Training Module Execution}    
		\label{algo1}
	}
\end{algorithm}

The proposed methodology (MAT-CNN-SOPC) is bio-inspired due to the fact that human beings constantly keep learning even when they are introduced to a completely new environment so that they could adjust to that environment quickly and adapt to it. By using this same concept we could enhance the learning mechanism of the CNN model for a particular scene-based application.

\subsection{Employed CNN Models}
In order to prove the effectiveness of our proposed methodology we chose two popular object-centric CNN architectures, VGG 16 convolutional-layer (VGG16) \cite{vgg2014} and ResNet50 \cite{resnet2016} CNN. The selected CNN architectures contain 138 million parameters for VGG16 and 26 million parameters for ResNet50. 

A typical approach to enable training of very deep networks on small datasets is to use a model pre-trained on a very large dataset, and then use the CNN as as an initialization for the task of interest. This method, referred to as `transfer learning' \footnote{~Transfer is achieved by taking the convolutional base of a pre-trained network, running the new data of 4 traffic categories through it and training a new, randomly initialized classifier on top of the semantic image output vector $\textbf{Y}_{out}$.} \cite{transferLearning2010, zeiler2014transferLearning} injects knowledge from other tasks by deploying weights and parameters from a pre-trained network to the new one. The rationale behind this is that the internal layers of the CNN can act as a generic extractor of image representations which have been pre-trained on one large dataset (source task) and then re-used on other target tasks. Considering the size of the dataset we have used (see Sec. \ref{datasetUsed}), the only way to apply a deep CNN such as VGG16 and ResNet50, is to reduce the number of trainable parameters. In order to achieve this the first filter stages are held fixed during training (weights are not updated) and overfitting \footnote{~Overfitting happens when the CNN model recognizes specific images in your training set instead of general patterns.} can be avoided. We initialize the feature extraction modules using pre-trained models from a large scale dataset, ImageNet \cite{alexnet2012, imagenetChallenge2015}. For the target task (traffic analysis), we design a network that will output scores for the 4 target categories of the dataset used.

\section{Experimental Results}\label{results}

\subsection{Experimental Setup}\label{expSetup}
For this research we have taken the 400 highway images (mentioned in Section \ref{datasetUsed}) and have used that for our training and validation purposes. The dataset is partitioned into two dataset consisting of training and validation sets and during every test randomization algorithm was used on the whole dataset to create the training and validation subsets. We have selected 3 random videos from each category (light, heavy and traffic) of the UCSD dataset and then converted the video stream to image by processing 1 frame out of every 8 frames (\textasciitilde1.3 \textit{fps}). Since the videos from the UCSD dataset is categorized based on light, heavy and traffic jams, we had to manually categorized into our generic 4 categories: Jam, Heavy, Fluid, Empty and generated 192 images (48 images for each category) for testing purposes. We have performed the following tests, which are separated into groups, as follows:

\renewcommand{\arraystretch}{1.4}
\renewcommand{\tabcolsep}{4pt}
\begin{table}[htbp]\label{testSetupTable}
	\caption{Tests Performed}
	\begin{center}
		\begin{tabular}{ |p{2cm}|p{2cm}|p{2cm}|  }
			\hline
			\multicolumn{3}{|c|}{\textbf{Test Groups}} \\
			\cline{1-3} 
			\textbf{G1: VGG16 performance on Dataset} & \textbf{G2: VGG16 performance on UCSD Dataset}& \textbf{\textit{G3: ResNet50 performance on UCSD Dataset}} \\
			\hline
			(i) 90\% Training / 10\% Validation& (i) 90\% Training / 10\% Validation & (i) 90\% Training / 10\% Validation   \\
			\hline
			(ii) 80\% Training / 20\% Validation& (ii) 75\% Training / 25\% Validation & (ii) 75\% Training / 25\% Validation \\
			\hline
			(iii) 70\% Training / 30\% Validation& (iii) 50\% Training / 50\% Validation& (iii) 50\% Training / 50\% Validation\\
			\hline
			(iv) 60\% Training / 40\% Validation& (iv) 75\%Training / 25\% Validation with reFeed Image Stack Feature& (iv) 75\%Training / 25\% Validation with reFeed Image Stack Feature \\
			\hline
		\end{tabular}
		\label{tab1}
	\end{center}
\end{table}

In Group 1 of tests (\textit{G1}), in test \textit{G1.i} we have broken the 400 training images into two dataset: 360 images for training and 40 images for validation (in 9:4 ratio) of VGG16 pre-trained model. In test \textit{G1.ii} we have broken the dataset into 320 for training and 80 for validation sets, whereas in \textit{G1.iii} it is broken in the ratio of 7:3 and in \textit{G1.iv} it is broken in 3:2. No separate tests were performed to check the accuracy of the categorization after training in Group 1 of tests, but the main motivation was to check the performance of training the VGG16 model on the 400 traffic images.

In Group 2 of tests we have taken the pre-trained VGG16 model and have trained the model with training and validation dataset in the ratio as mentioned in Table \ref{testSetupTable}. But in this group of tests we have checked the categorization accuracy of the model after training is complete with the 192 images of UCSD dataset as mentioned earlier in this section. The UCSD dataset was completely kept hidden during the training process so that we could evaluate the desirability of using VGG16 in scenarios of traffic analysis, which it has not been exposed to in advance (cross validation using unseen UCSD dataset). In Group 3 of tests we ran the similar set of tests as in Group 2 but we replaced the pre-trained CovNet model with ResNet50 and check the categorization accuracy with the UCSD dataset. For each test in every group, we have completely re-trained the CovNet model on our dataset to avoid bias of the model.

To prove our proposed MAT-CNN-SOPC model (Fig. \ref{mainModel}) and effective use of \textit{reFeed Image Stack} for further training (transfer learning), we have also performed a series of tests where the model is further trained with images from \textit{reFeed Image Stack}, which is segregated into training and validation set in the ratio of 75:25. Tests G2.iv and G3.iv represents those tests for VGG16 and ResNet50 respectively. To check the testing accuracy after this training method we used a different set of 192 images of the UCSD dataset for the purpose. We trained the CNN models for 10 epochs with a batch size of 10 images. Since we have worked with a small dataset for the problem in hand, we have used several image augmentation techniques such as Reflection \footnote{~Where each image is reflected horizontally.}, Translation \footnote{~Where each image is translated by a distance, measured in pixels.}, etc. to fit the training of the CNN model. We also implemented the training module on ZYNQ FPGA using Vivado HLS (see Section \ref{hardwaresoftware}). This is an alternate attempt to accelerate some of the functionalities of CNN.

\subsection{Classification Results}
For every single instance of the tests in each group (\textit{G1, G2, G3}) mentioned in the previous subsection (\ref{expSetup}), we have performed the same tests to check consistency and only the maximum result of those tests are reported in this section. In Table \ref{testResultTab} we could see the performance of each test, where validation accuracy along with categorization accuracy (testing) are reported.

\renewcommand{\arraystretch}{1.4}
\renewcommand{\tabcolsep}{4pt}
\begin{table}[htbp]\label{testResultTab}
	\caption{Tests Performed}
	\begin{center}
		\begin{tabular}{ |p{2cm}|p{2cm}|p{2cm}|  }
			\hline
			\multicolumn{3}{|c|}{\textbf{Results of Test Groups}} \\
			\cline{1-3} 
			\textbf{G1: VGG16 performance on Dataset} & \textbf{G2: VGG16 performance on UCSD Dataset}& \textbf{\textit{G3: ResNet50 performance on UCSD Dataset}} \\
			\hline
			(i) Validation Accuracy: 92.50\% & (i) Validation Accuracy: 90.00\%; \textbf{Testing Accuracy: 65.60}\% & (i) Validation Accuracy: 92.50\%; Testing Accuracy: 40.00\%   \\
			\hline
			(ii) Validation Accuracy: 87.50\% & (ii) Validation Accuracy: 89.50\%; Testing Accuracy: 60.00\% & (ii) \textbf{Validation Accuracy: 88.00\%; Testing Accuracy: 33.33}\%   \\
			\hline
			(iii) Validation Accuracy: 89.17\% & (iii) Validation Accuracy: 90.00\%; Testing Accuracy: 62.30\% & (iii) Validation Accuracy: 84.50\%; \textbf{Testing Accuracy: 61.67}\%   \\
			\hline
			(iv) Validation Accuracy: 89.38.50\%& (iv) \textbf{Validation Accuracy: 94.59\%; Testing Accuracy: 87.50}\% & (iv) \textbf{Validation Accuracy: 95.50\%; Testing Accuracy: 81.25}\% \\
			\hline
		\end{tabular}
		\label{tab1}
	\end{center}
\end{table}

 \begin{figure*}[!ht]\label{trainingGraph}
	\centering
	\subfigure[Result: Validation Accuracy \& Loss of VGG16 in G1.i Test]
	{\includegraphics[width=0.45\textwidth,height=6.5cm]{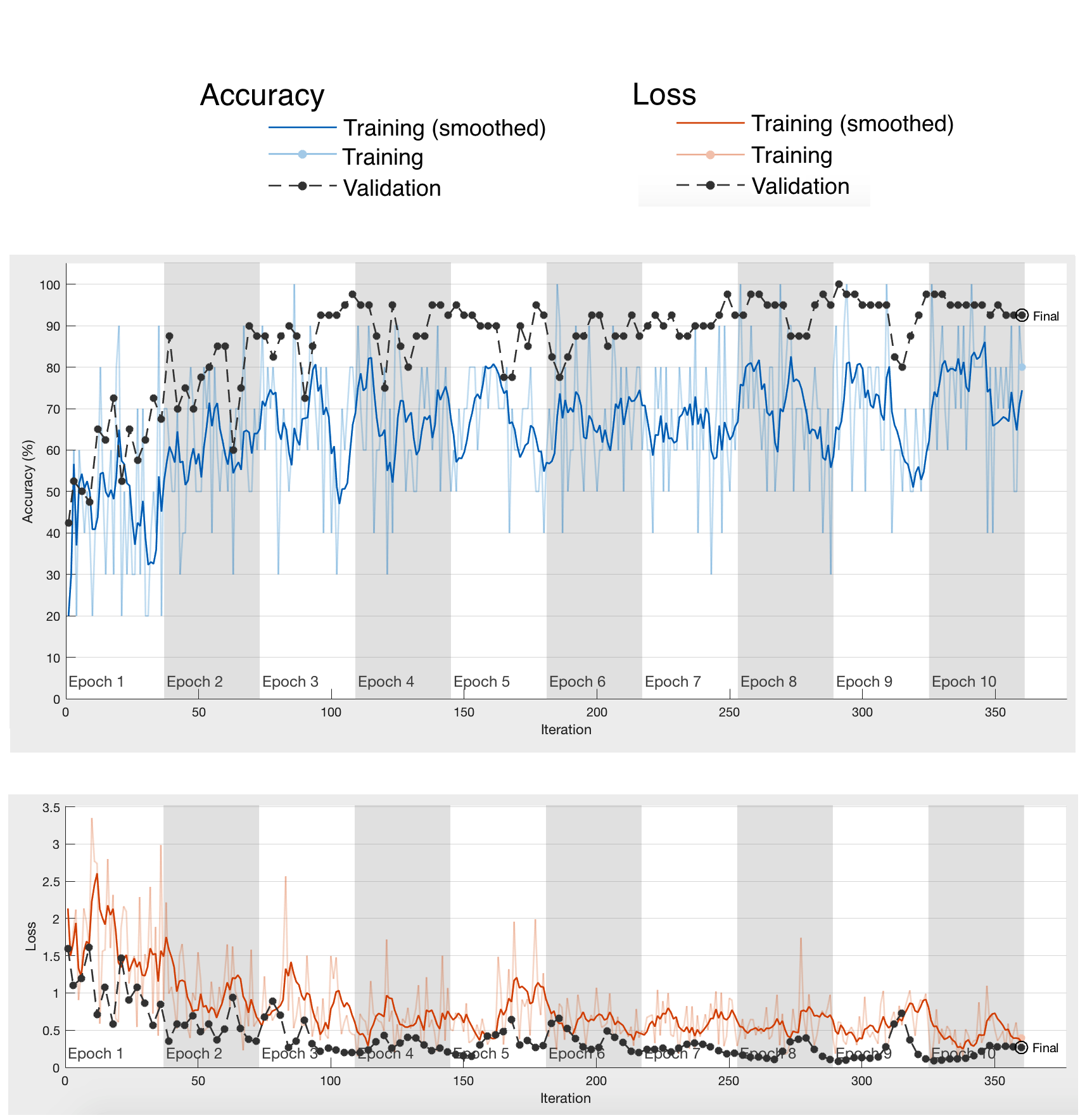}
		\label{fig01}}
	\subfigure[Result: Validation Accuracy \& Loss of ResNet50 in G3.i Test]
	{\includegraphics[width=0.46\textwidth,height=6.5cm]{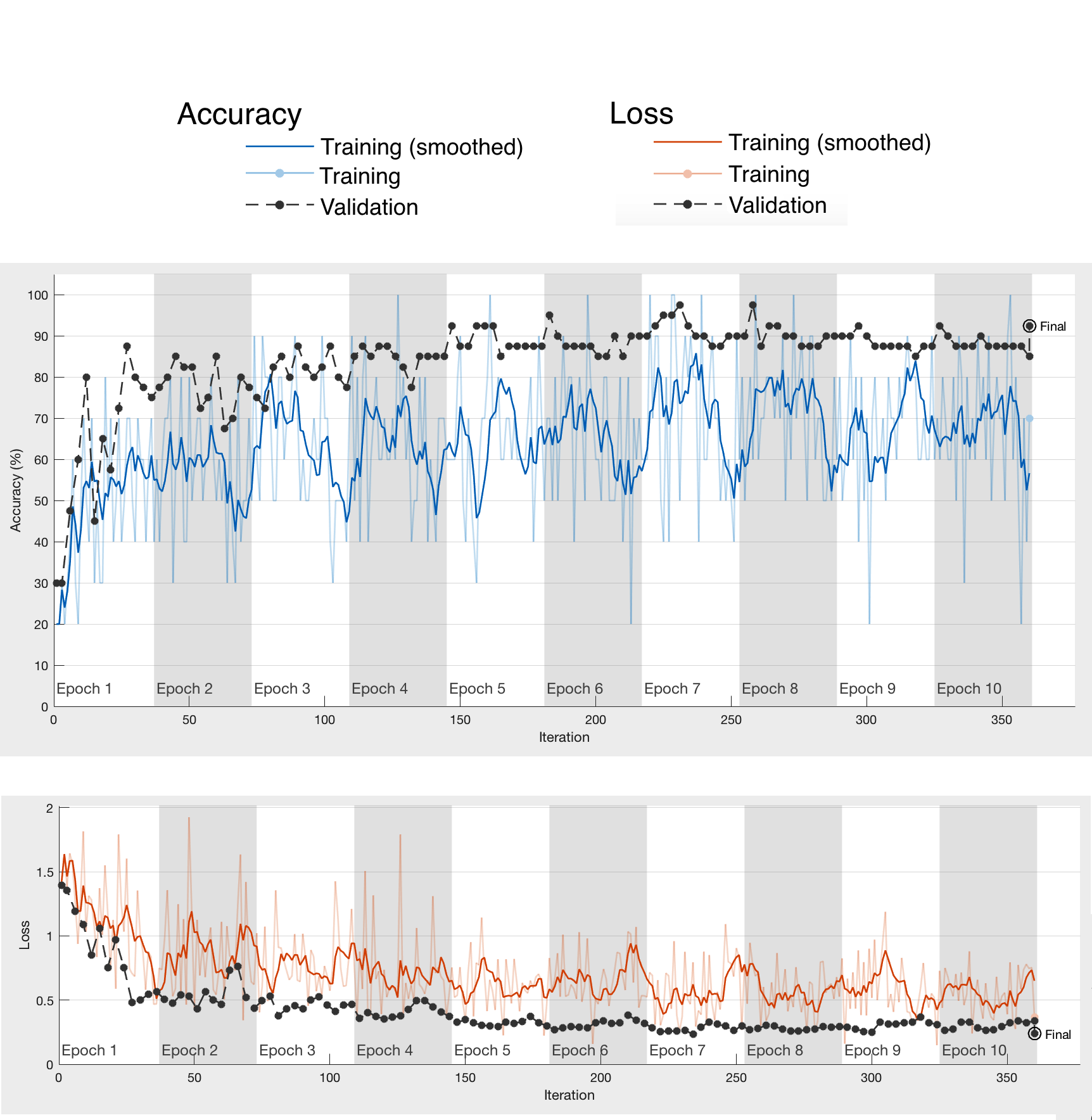}\label{fig02}}
	\caption{Graph Showing Validation Accuracy \& Loss}
\end{figure*}
\vspace{-0.5mm}

%

As we could see from Table \ref{testResultTab}, initially after using the stock traffic image dataset for training the testing prediction accuracy in \textit{G2.i} was 65.60\%, which was the highest in that group. But when we have used re-training mechanism (refer to Algo. \ref{algo1}) on the CNN model with \textit{re-Feed Image Stack}, the testing prediction accuracy got boosted to 87.50\% for the same group (\textit{G2}) and boosted to 81.25\% in \textit{G3} group compared to 33.33\% (without re-training). Although, it is a common knowledge that with more images for training accuracy of the CNN model improves but the images used for re-training did not exceed more than 10\% of the initial training (offline) dataset in size and given the size of the dataset we are working on, the gain (\textit{reFeed Gain}) in prediction accuracy is solely because of the methodology (training with \textit{reFeed Image Stack}) used rather than the possibility of using more images during training.

Now, using the Eq. \ref{eqrFG} and the resulting values from Table \ref{results}, the calculated \textit{reFeed Gain Factor} (\textit{r}) is 47.92 and the \textit{reFeed Gain} (\textit{R}) (using Eq. \ref{eqR}) is 2.44x for \textit{G3.iv}. Example 1 sheds some light on the phenomenon of enrichment of accuracy as described through \textit{reFeed Gain}.

\subsection*{Observation:}
\noindent{\textbf{Example 1.} In \textit{G3.iv}, the testing accuracy is 81.25\% ($P_{i}^{f}$), whereas in \textit{G3.iv} the testing accuracy is 33.33\% ($P_{i}^{0}$), thus from Eq. \ref{eqR}:}

\begin{equation*}
R = (81.25 / 33.33) = 2.4377 \approx 2.44
\end{equation*}

\noindent{Therefore, the boost in prediction accuracy for ResNet50 for this example using \textit{reFeed Image Stack} is 2.44x.}

 The hardware implementation is carried out on Zynq ZC-Z7045. It is observed that near about 95\% of DSP (858 out of 900), 55\% of BRAM (301 out of 545) and 41\% of LUTs (89626 out of 218600) have been utilized.

\section{Discussion}\label{discuss}
In the work \cite{luo2015traffic, luo2016traffic}, the authors have used the same 400 images dataset and have split it into two: Training and Testing, which means that the authors have used the same dataset for training, validation and testing, which is highly undesirable in this field to evaluate accuracy of the implemented CNN \footnote{~It is undesirable to use the same dataset for training, validation and testing since it introduces high level of bias.}. For example,  in \cite{luo2015traffic} they have used the same UCSD dataset to both train and test  the VGG model (after splitting the dataset into 75\% for training and 25\% for testing) and have achieved an accuracy of 96.10\%. This way of predicting accuracy of an application based CNN model is highly biased. When we trained our VGG 16 model with separate image dataset and tested the accuracy on the UCSD one, we got an accuracy of just 60.00\% (refer G2.ii in Tab. \ref{testResultTab}) in comparison. Additionally given the small size of the dataset used, there are two possible challenges, which could be faced. One of those issues being overfitting \footnote{~Overfitting happens when the CNN model recognizes specific images in your training set instead of general patterns.} The other issue is that the model might not be able to train properly and result into less accurate predictions. In \cite{luo2015traffic}, the reported accuracy results of the implemented models were on validation instead of reporting the testing accuracy of the same. When the UCSD dataset was used for testing and the curated 400 traffic images for training in our model, we found out that the testing accuracy was very less compared to the validation one, contradicting their results. In order to improve the testing accuracy of CNN models for traffic analysis we came up with MAT-CNN-SOPC Model. 

In Table \ref{results}, we could also see an anomaly in using ResNet50, where with less training images it performed better. One of the possible reasons being overfitting of images when trained with less number of images but from the training graphs (see Fig. \ref{trainingGraph}) we could understand that is not the case. The other possible reason being mislabeling of the images while testing. For our example we have noticed that sometimes it was difficult even for a human to differentiate between `Heavy' and `Fluid' traffic and since the testing images were labeled manually.

From the graphs in Fig. 3 we could also see that the model is somehow underfitting rather than overfitting, but incorporating the MAT-CNN-SOPC Model for the training and prediction has actually made the gap between the training, validation and testing accuracy narrower. Although it could be argued upon that since we have used images from the same camera and on the same road junction to improve the training quality of the CNN model but given the practical application of traffic analysis it is highly likely that the same camera system would operate in the same junction/street region for its lifetime. Thus training the camera system with known environment seems to solve the problem of analyzing and categorizing traffic in a cost effective way. Another noteworthy  thing to mention is that for this application and for our tests we have chosen the value of Quality of Experience (\textit{QoE}) as 70\% \footnote{~For our traffic categorization issue we found out through testing that chosing QoE value of 70\% produced better result in re-training the model for accuracy.} by default but, this value could be modified based on the desired accuracy for the problem in hand and we could also utilize Eq. \ref{eqRRelationship} to fine tune MAT-CNN-SOPC for the same purpose. 

\section{Conclusion}\label{conclusion}
In this paper, we have proposed a novel CNN based categorization model, which could categorize traffic effectively on the programmable system board even with less number of training images in the dataset. To effectively train the CNN to improve prediction accuracy, we have used a combination of transfer learning as well as a novel re-training mechanism on pre-trained CNN models, where the model is re-trained with images from a known environment. We have also introduced Quality of Experience, which researchers in this field could use to choose the right CNN model for their problem and achieve the desired results (in terms of accuracy).

\section*{Acknowledgment}
This work is supported by the UK Engineering and Physical Sciences Research Council EPSRC [EP/R02572X/1 and EP/P017487/1] and the authors would like to thank the people associated with National Centre for Nuclear Robotics (NCNR) and Extreme Environments for their support and feedback. Somdip would also like to thank everyone from the Embedded and Intelligent Systems Laboratory at the University of Essex for their feedback on this project. 

\bibliographystyle{IEEEtran}

\bibliography{research}

\end{document}